%% file: MAIN.tex
\documentclass[letterpaper, 10 pt, conference]{class/ieeeconf}  

\IEEEoverridecommandlockouts                              
\usepackage{xcolor}
\usepackage{siunitx}  
\usepackage{graphicx}
 \graphicspath{./figures/robot_demonstration}

\usepackage{amsmath}
\usepackage{amssymb}
\usepackage{latexsym}
\usepackage{url}
\usepackage{cite}
\usepackage{relsize}
\usepackage{multirow}
\usepackage{afterpage}
\usepackage{ifthen}
\usepackage{graphicx}
\usepackage{algpseudocode,algorithm,algorithmicx}
\usepackage{subfigure}
\usepackage{flushend}
\usepackage{epstopdf}
\usepackage{stackengine}
\usepackage{textcomp} 

\usepackage{gensymb}
\usepackage{booktabs}
\usepackage{makecell}
\usepackage{hhline}
\usepackage{courier}
\usepackage{lipsum}
\usepackage{xspace}
\usepackage{xcolor} 
            
\makeatletter
\let\NAT@parse\undefined
\makeatother
\usepackage[bookmarks=false, linkcolor=blue, urlcolor=blue, citecolor=blue]{hyperref} 
\hypersetup{
    colorlinks=true,
    linkcolor=red,
    filecolor=magenta,      
    urlcolor=blue,
    pdfstartview={FitH},
    citecolor =blue
    }

\usepackage{xspace}

\newcommand{\etal}{\textit{et al.}}

 \graphicspath{{./figures/}}

\title{\LARGE \bf GraspMamba: A Mamba-based Language-driven Grasp Detection Framework with Hierarchical Feature Learning}

\author{Huy Hoang Nguyen$^{1,2}$, An Vuong$^{3}$, Anh Nguyen$^4$, Ian Reid$^{3}$, Minh Nhat Vu$^{1,2}$ 
\thanks{$^1$ Automation \& Control Institute, TU Wien, Austria 
}
\thanks{$^2$ Austrian Institute of Technology (AIT) GmbH, Austria  
}
\thanks{$^3$ Department of Computer Vision, MBZUAI, UAE 
}
\thanks{$^4$ Department of Computer Science, University of Liverpool, UK 
}
}

\begin{document}
\input{0_localmacros}

\maketitle
\thispagestyle{empty}
\pagestyle{empty}

\begin{abstract}
Grasp detection is a fundamental robotic task critical to the success of many industrial applications. However, current language-driven models for this task often struggle with cluttered images, lengthy textual descriptions, or slow inference speed. We introduce GraspMamba, a new language-driven grasp detection method that employs hierarchical feature fusion with Mamba vision to tackle these challenges. By leveraging rich visual features of the Mamba-based backbone alongside textual information, our approach effectively enhances the fusion of multimodal features. GraspMamba represents the first Mamba-based grasp detection model to extract vision and language features at multiple scales, delivering robust performance and rapid inference time. Intensive experiments show that GraspMamba outperforms recent methods by a clear margin. We validate our approach through real-world robotic experiments, highlighting its fast inference speed.
\end{abstract}


\input{1_intro}

\input{2_rw}

\input{3_method}

\input{4_exp}

\input{5_conclusions}

\bibliographystyle{class/IEEEtran}
\bibliography{class/IEEEabrv,class/reference}
   
\end{document}

%% file: 0_localmacros.tex

\newtheorem{problem}{Problem}
\newtheorem{lemma}{Lemma}
\newtheorem{theorem}[lemma]{Theorem}
\newtheorem{claim}{Claim}
\newtheorem{corollary}[lemma]{Corollary}
\newtheorem{definition}[lemma]{Definition}
\newtheorem{proposition}[lemma]{Proposition}
\newtheorem{remark}[lemma]{Remark}
\newenvironment{LabeledProof}[1]{\noindent{\it Proof of #1: }}{\qed}

\def\beq#1\eeq{\begin{equation}#1\end{equation}}
\def\bea#1\eea{\begin{align}#1\end{align}}
\def\beg#1\eeg{\begin{gather}#1\end{gather}}
\def\beqs#1\eeqs{\begin{equation*}#1\end{equation*}}
\def\beas#1\eeas{\begin{align*}#1\end{align*}}
\def\begs#1\eegs{\begin{gather*}#1\end{gather*}}

\newcommand{\poly}{\mathrm{poly}}
\newcommand{\eps}{\epsilon}
\newcommand{\e}{\epsilon}
\newcommand{\polylog}{\mathrm{polylog}}
\newcommand{\rob}[1]{\left( #1 \right)} 
\newcommand{\sqb}[1]{\left[ #1 \right]} 
\newcommand{\cub}[1]{\left\{ #1 \right\} } 
\newcommand{\rb}[1]{\left( #1 \right)} 
\newcommand{\abs}[1]{\left| #1 \right|} 
\newcommand{\zo}{\{0, 1\}}
\newcommand{\zonzo}{\zo^n \to \zo}
\newcommand{\zokzo}{\zo^k \to \zo}
\newcommand{\zot}{\{0,1,2\}}
\newcommand{\en}[1]{\marginpar{\textbf{#1}}}
\newcommand{\efn}[1]{\footnote{\textbf{#1}}}
\newcommand{\vecbm}[1]{\boldmath{#1}} 
\newcommand{\uvec}[1]{\hat{\vec{#1}}}
\newcommand{\thv}{\vecbm{\theta}}
\newcommand{\junk}[1]{}
\newcommand{\var}{\mathop{\mathrm{var}}}
\newcommand{\rank}{\mathop{\mathrm{rank}}}
\newcommand{\diag}{\mathop{\mathrm{diag}}}
\newcommand{\tr}{\mathop{\mathrm{tr}}}
\newcommand{\acos}{\mathop{\mathrm{acos}}}
\newcommand{\atantwo}{\mathop{\mathrm{atan2}}}
\newcommand{\SVD}{\mathop{\mathrm{SVD}}}
\newcommand{\quadf}{\mathop{\mathrm{q}}}
\newcommand{\linterp}{\mathop{\mathrm{l}}}
\newcommand{\sgn}{\mathop{\mathrm{sign}}}
\newcommand{\sym}{\mathop{\mathrm{sym}}}
\newcommand{\avg}{\mathop{\mathrm{avg}}}
\newcommand{\mean}{\mathop{\mathrm{mean}}}
\newcommand{\erf}{\mathop{\mathrm{erf}}}
\newcommand{\grad}{\nabla}
\newcommand{\R}{\mathbb{R}}
\newcommand{\defeq}{\triangleq}
\newcommand{\dims}[2]{[#1\!\times\!#2]}
\newcommand{\sdims}[2]{\mathsmaller{#1\!\times\!#2}}
\newcommand{\udims}[3]{#1}
\newcommand{\udimst}[4]{#1}
\newcommand{\com}[1]{\rhd\text{\emph{#1}}}
\newcommand{\ind}{\hspace{1em}}
\newcommand{\argmin}[1]{\underset{#1}{\operatorname{argmin}}}
\newcommand{\floor}[1]{\left\lfloor{#1}\right\rfloor}
\newcommand{\step}[1]{\vspace{0.5em}\noindent{#1}}
\newcommand{\quat}[1]{\ensuremath{\mathring{\mathbf{#1}}}}
\newcommand{\norm}[1]{\left\lVert#1\right\rVert}
\newcommand{\ignore}[1]{}
\newcommand{\specialcell}[2][c]{\begin{tabular}[#1]{@{}c@{}}#2\end{tabular}}
\newcommand*\Let[2]{\State #1 $\gets$ #2}
\newcommand{\algorithmicbreak}{\textbf{break}}
\newcommand{\Break}{\State \algorithmicbreak}
\newcommand{\ra}[1]{\renewcommand{\arraystretch}{#1}}

\renewcommand{\vec}[1]{\mathbf{#1}} 

\algdef{S}[FOR]{ForEach}[1]{\algorithmicforeach\ #1\ \algorithmicdo}
\algnewcommand\algorithmicforeach{\textbf{for each}}
\algrenewcommand\algorithmicrequire{\textbf{Require:}}
\algrenewcommand\algorithmicensure{\textbf{Ensure:}}
\algnewcommand\algorithmicinput{\textbf{Input:}}
\algnewcommand\INPUT{\item[\algorithmicinput]}
\algnewcommand\algorithmicoutput{\textbf{Output:}}
\algnewcommand\OUTPUT{\item[\algorithmicoutput]}

%% file: 1_intro.tex
\section{INTRODUCTION} \label{Sec:Intro}
Robotic grasping is an important task with several applications and has received growing attention from researchers in the past decades~\cite{sundermeyer2021contact,wen2022catgrasplearning,nguyen2016preparatory,ainetter2022endtoend}. Traditional grasp detection methods~\cite{redmon2015realtime} often overlook the use of language prompts~\cite{gilles2023metagraspnetv2}, limiting the agent’s ability to interpret language instructions beyond their literal meaning and resulting in unpredictable behavior~\cite{schirmer2024towards}. Emerging language-driven grasp methodologies~\cite{vuong2023grasp, yuan2023m2t2} have enabled robots to grasp specific objects based on language prompts~\cite{vuong2024language}. These robotic systems, trained on large-scale datasets, provide robust visual-semantic understanding, like determining which object part to grasp based on human instructions~\cite{yuan2024robopoint}, showing promise in developing general-purpose robots~\cite{o2024open}.

Recently, with the rise of large vision language models, language-driven grasping has gained attention as a promising research area in robotic manipulation. The use of language can be viewed as a representation of the scenario surrounding objects, conveying information to humans or machines for conceptual understanding. For example, by giving a command to ``grasp a cup on the table," the robot knows where a cup is and can determine the specific grasp actions for objects. Therefore, by leveraging language, many studies attempt to bridge the gap between vision and language for robotic applications~\cite{vanvo2024languagedrivengraspdetectionmaskguided,nghia2024fastgrasp,vuong2024language}. For example, SayCan~\cite{ahn2022icanisay} and PaLM-E~\cite{driess2023palm} are robotic language models designed to provide instructions for robots operating in real-world environments by leveraging large language models such as~\cite{chatgpt} or large vision language models~\cite{shridhar2022cliport}. 

Previous studies have explored various methods to address the challenge of integrating language in grasp detection task. One approach treats this as a grasp-pose generation problem conditioned on language prompts~\cite{xu2023jointmodelingvision} or uses diffusion models~\cite{vuong2024language} to achieve promising results. However, diffusion models face challenges in real-time robotics due to their long inference times~\cite{nghia2024fastgrasp}. Alternatively, methods that leverage transformers~\cite{shridhar2022cliport, mirjalili2023langraspusinglargelanguage, vanvo2024languagedrivengraspdetectionmaskguided} successfully combine textual and visual features but often struggle with object complexities~\cite{tang2023graspgpt}. This limitation is particularly evident in scenarios requiring long-range visual-language dependencies and handling images with extensive textual descriptions, which restricts their application to real-world robots.

Recently, Mamba~\cite{gu2024mamba}, with its global receptive field coverage and dynamic weights offering linear complexity~\cite{han2024mamba3d}, presents an ideal solution for addressing long-range visual-language dependencies while maintaining fast inference speeds, making it well-suited for robotic applications~\cite{liu2024robomamba}. Mamba has demonstrated exceptional effectiveness in tasks involving long sequence modeling, especially in natural language processing~\cite{lieber2024jamba}. Researchers have begun exploring its potential for vision-related applications, including image classification~\cite{chen2024resvmamba,chen2024rsmambaremotesensingimage,wang2024s2mambaspatialspectralstatespace}, image segmentation~\cite{ma2024rs3mambavisualstatespace,zhu2024samba,wan2024sigmasiamesemambanetwork}, and point cloud analysis~\cite{liang2024pointmamba,wang2024pointramba,han2024mamba3d}. Roboticists are also investigating how Mamba's context-aware reasoning and linear complexity can be applied to solve robotic tasks~\cite{liu2024robomamba, jia2024mail}. In line with this direction, this paper aims to leverage Mamba to integrate image and text modalities to generate semantically plausible grasping poses for robotic systems.

This paper introduces GraspMamba, the first language-driven grasp detection framework built on the State Space Model. Specifically, we propose a novel hierarchical feature fusion technique that integrates textual features with visual features at each stage of the hierarchical vision backbone. We argue that our Mamba-based fusion technique addresses the computational inefficiencies of transformers caused by the doubled sequence length when combining text and vision features~\cite{fang2019scene}. Our method maintains strong performance in grasp detection by efficiently learning spatial information and hierarchically incorporating textual features to enhance context and semantics. By aligning textual features within a shared space, the model effectively merges multimodal representations across multiple scales. We validate our approach on a recent large-scale language-driven grasping dataset~\cite{vuong2024language}, demonstrating better accuracy and faster inference than current state-of-the-art methods. Additionally, our approach supports zero-shot learning and is generalized to real-world robotic grasping applications.


Our main contributions are summarized as follows:
\begin{itemize}
    \item We propose a first vision-language model based on the Mamba technique aiming to fuse vision-language features in a hidden state space, which opens up a new paradigm for cross-modality feature fusion for the language-driven grasp detection task.
    \item We provide an in-depth analysis of our proposed method, presenting experimental results on benchmark datasets. Our findings demonstrate that it surpasses other approaches in accuracy and execution speed. Our code and models will be released.
\end{itemize}

%% file: 2_rw.tex
\section{Related Work} \label{Sec:rw}
\textbf{Grasp Detection.} Traditional robot grasp detection methods include analytic approaches~\cite{domae2014fast,bohg2014data} that rely on kinematic and dynamic models to identify stable grasp points, ensuring they meet flexibility, balance, and stability criteria. In contrast, several data-driven approaches based on machine learning~\cite{zhou2018fullyconvolutional,yu2023egnet,guo2017ahybrid} have been developed to enable robots to learn and mimic human grasping strategies through deep learning techniques. These approaches have been further enhanced by the use of RGB-D images~\cite{yu2022seresunet,tong2024anovelrgbd} and 3D point clouds~\cite{wang2021graspness,ni2020pointnet,nguyen2023open,nguyen2024LGrasp6D}, allowing for grasp detection in 3D space. However, a major limitation of both analytic and CNN-based methods is their restricted scene understanding and inability to process language instructions, which reduces their effectiveness in dynamic, human-centered environments.

\textbf{Language-driven Grasping}. Language-driven grasping represents the use of natural language to localize the object region for grasping~\cite{lu2023vl, sun2023language, nguyen2024LGrasp6D, cheang2022learning, vuong2024language,nghia2024fastgrasp, xu2023joint, yang2022interactive}. Recent research focuses on methods that establish the correlation between textual embeddings and vision embeddings within the embedding space. This approach aims to identify the target object and subsequently generate the grasping pose based on the combined features~\cite{vuong2023grasp}. Jin~\etal~\cite{jin2024reasoning} introduces a method that leverages the reasoning capabilities of a large language model to generate grasping poses based on indirect verbal instructions. However, a significant limitation of these methods is their high computational and memory demands during training and inference are a significant limitation.

\textbf{Cross-Modal Feature Fusion.} 
Multimodal feature fusion is a critical technique in various applications to optimize the alignment between linguistic and visual domains. Conventionally, many approaches have focused on independently mapping global features of images and sentences into a shared embedding space to compute image-sentence similarity~\cite{huang2016instanceaware,huang2017learning}. Recent advancements, such as the work by Xu~\etal~\cite{xu2021crossmodal}, capture higher-order interactions between visual regions and textual elements by incorporating inter- and intra-modality relations in the feature fusion process. Furthermore, the attention mechanism introduced in the Transformer~\cite{vaswani2017attention} and cosine similarity metrics have demonstrated effectively between textual and visual embeddings. However, these multimodal feature fusion methods face limitations when they need to focus on fine-grained image regions or when processing queries with limited textual information or complex sentences. 


\textbf{State Space Models.} State Space Models (SSMs) with selection mechanisms and hardware-aware architectures have recently demonstrated substantial promise in long-sequence modeling. The original SSM block is designed for processing one-dimensional sequences, while vision-related tasks necessitate handling multi-dimensional inputs like images, videos, and 3D representations. Several approaches have been proposed to adapt SSMs for complex vision-related applications. For instance, ViM~\cite{zhu2024visionmamba}, also called the Bidirectional Mamba block, annotates image sequences with position embeddings and condenses visual representations using bidirectional state space models.
Additionally, PlainMamba~\cite{yang2024plainmamba} and EfficientVMamba~\cite{pei2024efficientvmamba} improve the capabilities of visual state space~\cite{liu2024vmamba} blocks by stacking multiple blocks on the feature map and employing different scanning approaches. While these methods effectively address the need for global context and spatial understanding, their increased complexity can lead to challenges in training and a higher risk of overfitting. To mitigate these issues, Hatamizadeh~\etal~\cite{hatamizadeh2024mambavision} introduced a hybrid Mamba-Transformer backbone that enhances global context representation learning.

Despite Mamba’s increasing popularity in vision tasks~\cite{liang2024pointmamba,zhao2024cobra}, there remains a significant gap in integrating text and image modalities~\cite{qiao2024vlmamba}. A key challenge in vision-and-language Mamba models is using a single projection from the image to the language domain~\cite{xu2024survey}, which fails to capture image features at multiple resolutions~\cite{liu2021swin}. To mitigate this issue, we propose a hierarchical feature fusion method that integrates vision and text features at various scales, leveraging Mamba’s efficient computation~\cite{hatamizadeh2024mambavision}. Specifically, we integrate rich textual information from a text encoder at each stage of the vision backbone to enhance global multimodal information, retaining all crucial features through an element-wise technique. Consequently, the output for grasp detection preserves the essential information from the input data at multiple scales, which can serve as robust guidance to solve such fine-grained generative problem~\cite{zhong2023multi} like the language-driven grasp generation. Experimental results confirm that our Mamba-based fusion method delivers competitive performance with faster, linearly scalable inference and constant memory usage in both vision and robotic applications.

%% file: 3_method.tex
\section{GraspMamba} \label{Sec:method}

\begin{figure*}[!ht]
	\centering	
	\includegraphics[width=1.0\linewidth, trim=0 0 0.8cm 0, clip]{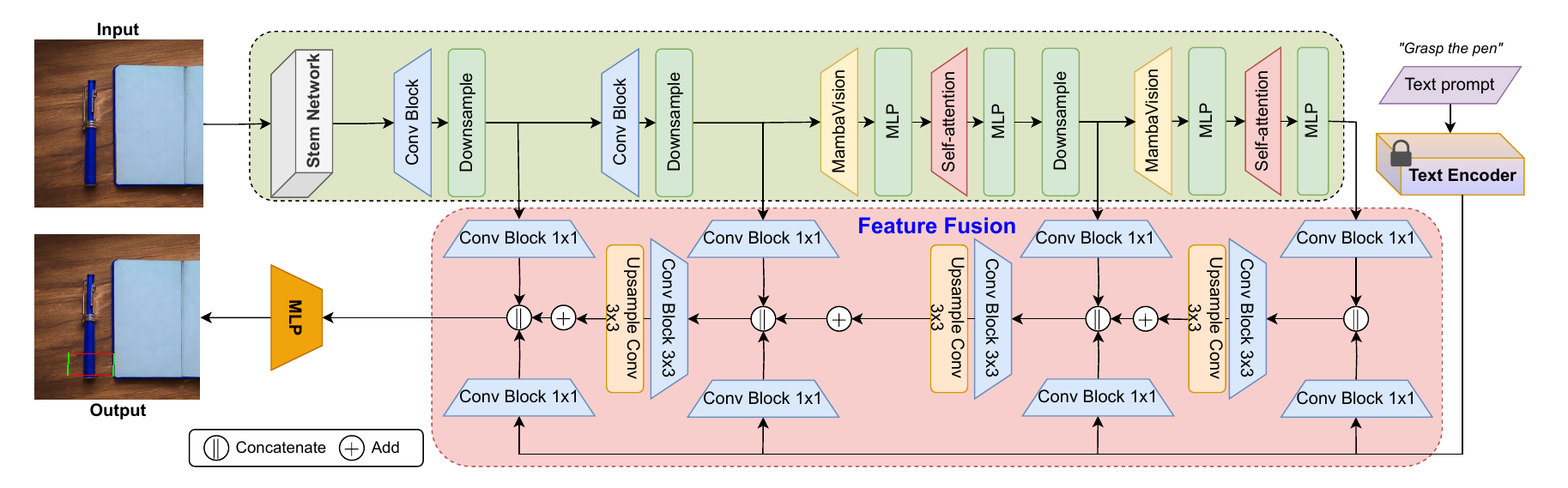}
	\vspace{1pt}
	\caption{The overview of our GraspMamba framework for the language-driven grasp detection task.}
	\label{fig:architecture}
\end{figure*}

\subsection{Overview}
We propose a method for detecting the grasping pose of an object by integrating textual features with rich visual features derived from a Mamba-based architecture. Given an input RGB image and a corresponding text prompt describing the object of interest, our approach aims to identify the object's grasping pose accurately. Following the established convention of \textit{rectangle grasp}, as outlined in \cite{depierre2018jacquard}, we define each grasping pose using five parameters: the center coordinates $(x, y)$, the rectangle's width and height $(w, h)$, and the rotational angle that indicates the rectangle's orientation relative to the image's horizontal axis. The overall framework of our method is depicted in Fig.~\ref{fig:architecture}. 

\subsection{Visual and Language Feature Extraction}

\textbf{Mamba Visual Feature Extraction.} Inspired by the powerful Swin Transformer's~\cite{liu2021swin} hierarchical design, we also adopts a four-stage structure to balance speed and accuracy in vision tasks~\cite{hatamizadeh2024mambavision}. Therefore, we leverage pre-trained Mamba vision weights to produce multi-level representations at each stage. Given an image of size $H \times W \times 3$, the initial two stages consist of CNN-based layers for fast feature extraction at higher input resolutions with size of $\frac{H}{4} \times \frac{W}{4} \times C$ and $\frac{H}{8} \times \frac{W}{8} \times 2C$, respectively. Subsequently, the MambaVision and multihead self-attention~\cite{vaswani2017attention} blocks are applied afterward as referred to as feature transformation stages, with output resolution of $\frac{H}{16} \times \frac{W}{16}$ and $\frac{H}{32} \times \frac{W}{32}$, respectively. Specifically, the MambaVision block modifies the original Mamba by creating the symmetric path without SSM as a token mixer to enhance the modeling of the global context. SSMs map one-dimensional sequence $\mathbf{x}(t) \in \mathbb{R}^L$ to $\mathbf{y}(t) \in \mathbb{R}^L$ through a hidden state $\mathbf{h}(t) \in \mathbb{R}^N$. With the evolution parameter $\mathbf{A} \in \mathbb{R}^{N \times N}$ and the projection parameters $\mathbf{B} \in \mathbb{R}^{N \times 1}$, $\mathbf{C} \in \mathbb{R}^{1 \times N}$, such a model is formulated as linear ordinary differential equations:
\begin{subequations}\label{eq: linear ODE}
\begin{align}
    \label{eq: linear ODE 1}
    \mathbf{h}'(t) &= \mathbf{A}\mathbf{h}(t) + \mathbf{B}\mathbf{x}(t), \\
    \label{eq: linear ODE 2}
    \mathbf{y}(t) &= \mathbf{C}\mathbf{h}(t).
\end{align}
\end{subequations}

As continuous-time models, state space models are adapted for deep learning applications with discrete data space through a discretization step using Zero-Order Hold assumption~\cite{gu2024mamba}. In this transformation, the continuous-time parameters $\mathbf{A}$ and $\mathbf{B}$ are converted into their discrete-time equivalents, denoted as $\overline{\mathbf{A}}$ and $\overline{\mathbf{B}}$, respectively, with a timescale parameter $\Delta$ according to:
\begin{subequations}\label{eq: linear matrices}
\begin{align}
    \label{eq: linear matrices 1}
    \overline{\mathbf{A}} &= \exp(\Delta\mathbf{A}), \\
    \label{eq: linear matrices 2}
    \overline{\mathbf{B}} &= (\Delta\mathbf{A})^{-1}(\exp(\Delta\mathbf{A}) - \mathbf{I}) \cdot \Delta\mathbf{B}.
\end{align}
\end{subequations}

Thus, Equation~\eqref{eq: linear ODE} can be rewritten as:
\begin{subequations}\label{eq: iterative equation}
\begin{align}
    \label{eq: iterative equation 1}
    \mathbf{h}_t &= \overline{\mathbf{A}}\mathbf{h}_{t-1} + \overline{\mathbf{B}}\mathbf{x}_t, \\
    \label{eq: iterative equation 2}
    \mathbf{y}_t &= \mathbf{C}\mathbf{h}_t.
\end{align}
\end{subequations}

To improve computational performance and allow for better scaling, the iterative process in Equation~\eqref{eq: iterative equation} can be synthesized through a global convolution
\begin{subequations}
\begin{align}
    \overline{\mathbf{K}} &= (\mathbf{C}\overline{\mathbf{B}}, \mathbf{C}\overline{\mathbf{A}}\overline{\mathbf{B}}, \cdots, \mathbf{C}\overline{\mathbf{A}}^{L-1}\overline{\mathbf{B}}), \\
    \mathbf{y} &= \mathbf{x} \ast \overline{\mathbf{K}},
\end{align}
\end{subequations}
where $L$ is the length of the input sequence $\mathbf{x}$, $\overline{\mathbf{K}} \in \mathbb{R}^L$ serves as the kernel of the SSMs and $\ast$ represents the convolution operation. As a result, by leveraging hierarchical representations from each stage of the Mamba-based backbone, we eagerly captures both large-scale structures and fine-grained details, enhancing its overall performance in various vision-related tasks.

\textbf{Text embedding.} Following the standard practice~\cite{vuong2024language}, we encode the input query (e.g., ``grasp a pencil") using a pre-trained BERT~\cite{devlin2018bert} or CLIP~\cite{radford2021learning} model, producing text embedding features $\mathbf{T} \in \mathbb{R}^{B \times C_\mathbf{T}}$.

\subsection{Hierarchical Feature Fusion}
While Mamba is effective at modeling long sequences, many Mamba-based multimodal approaches treat multimodal data as a single-domain sequence rather than concentrating on how to integrate features effectively~\cite{li2024mambadfusel,li2024coupledmamba,zhao2024cobra}. To address this, an inspired by hierarchical nature of Swin Transformer~\cite{liu2021swin} we aim to develop a new approach to fuse visual and textual features in a multi-scale manner. Additionally, our hierarchical feature fusion is a simple learnable module, aligns the vision and text features by transforming the dimensionality of the textual representation to match the token dimensions used in the Mamba-based model to create rich, multi-modal representation for tasks such as visual grounding or language-driven grasp detection. Our hierarchical feature fusion block is shown Fig.~\ref{fig:architecture}.


To effectively combine visual and textual information, we begin by aligning their dimensions and preparing them for fusion. This process involves applying $1 \times 1$ convolutions to both image and text features, which serves to reduce their channel dimensions to a common space while allowing the network to learn combined feature representations for fusion. We then expand the text features to match the spatial dimensions of the image features, ensuring that each spatial location in the image can attend to the entire text representation. These processed features are then concatenated along the channel dimension, creating a unified representation that preserves information from both modalities. Let \( \mathbf{X}_l \in \mathbb{R}^{B \times C_I \times H_l \times W_l} \) represent the image features at level \( l \in \{1, \dots, L\} \), \( \mathbf{T} \in \mathbb{R}^{B \times C_\mathbf{T}} \) represent the text features, and \( \mathbf{T}_{\text{exp}} \) represents the expanded text features to match the spatial dimensions of \( \mathbf{X}_l \). The visual-language fusion at level \( l \), denoted as \( \Phi_l \), is defined as:

\begin{equation}
   Z_l = \text{Concat}\left( \text{Conv}_{1 \times 1}^\mathbf{X}(\mathbf{X}_l), \text{Conv}_{1 \times 1}^\mathbf{T}(\mathbf{T}_{\text{exp}}) \right)
\end{equation}

To further integrate the concatenated features and capture spatial relationships, we apply a 3x3 convolution. This crucial step allows for local feature interactions between the image and text modalities, helps in learning spatially-aware multi-modal representations, and increases the receptive field to capture more context. This integration is achieved through:
\begin{equation}
   \Phi_l(\mathbf{X}_l, \mathbf{T}) = \text{Conv}_{3 \times 3} \left( Z_l \right) 
\end{equation}
where \( \text{Conv}_{1 \times 1}^\mathbf{X} \) and \( \text{Conv}_{1 \times 1}^\mathbf{T} \) are \( 1 \times 1 \) convolutions applied to the image and text features, respectively. \( \text{Conv}_{3 \times 3} \) is a \( 3 \times 3 \) convolution. 

To preserve global information across different levels of the vision backbone, we define an upscaling operation \( U_l \), which allows information to flow from deeper layers to shallower layers, helps to preserve fine-grained details while incorporating global context, and enables the model to make more informed decisions at each level of the hierarchy. The upscaling operation, applied to the hierarchical feature fusion \( F_l \) at layer \( l \) is defined as:

\begin{equation}
    U_l(F_l) = \text{BilinearUpsample}\left( \text{Conv}_{3 \times 3}(F_l) \right)
\end{equation}

Here, \( F_l \) represents the hierarchical feature fusion at layer \( l \), which is introduced recursively to capture and refine multi-scale information across the network. This recursive property is crucial as it allows for progressive refinement of features from the deepest to the shallowest layers, enables the model to capture multi-scale information effectively, and facilitates the integration of global and local features at each level. The hierarchical feature fusion is recursively defined as:

\begin{equation}
    F_l =
\begin{cases}
\Phi_L(\mathbf{X}_L, \mathbf{T}), & \text{if } l = L, \\
\Phi_l(\mathbf{X}_l, \mathbf{T}) + U_{l+1}(F_{l+1}), & \text{if } 1 \leq l < L.
\end{cases}
\end{equation}

Inspired by GR-ConvNet~\cite{kumra2020antipodal}, the final high-dimensional features $F_l$ are subsequently transformed into the grasp detection output through a composition of multiple MLP layers.



%% file: 4_exp.tex
\section{Experimental Results} 
\label{Sec:exp}
The experiments initially focus on evaluating the effectiveness of our approach on the Grasp-Anything dataset~\cite{vuong2023grasp}. Following this, we test our proposed method on real robot grasp detection tasks. Additionally, we conduct ablation studies to analyze our method in the context of language-driven grasp detection. Finally, we discuss the challenges faced and highlight open questions for future research.

\begin{figure*}[t]
	\centering
	\includegraphics[width=1.0\linewidth]{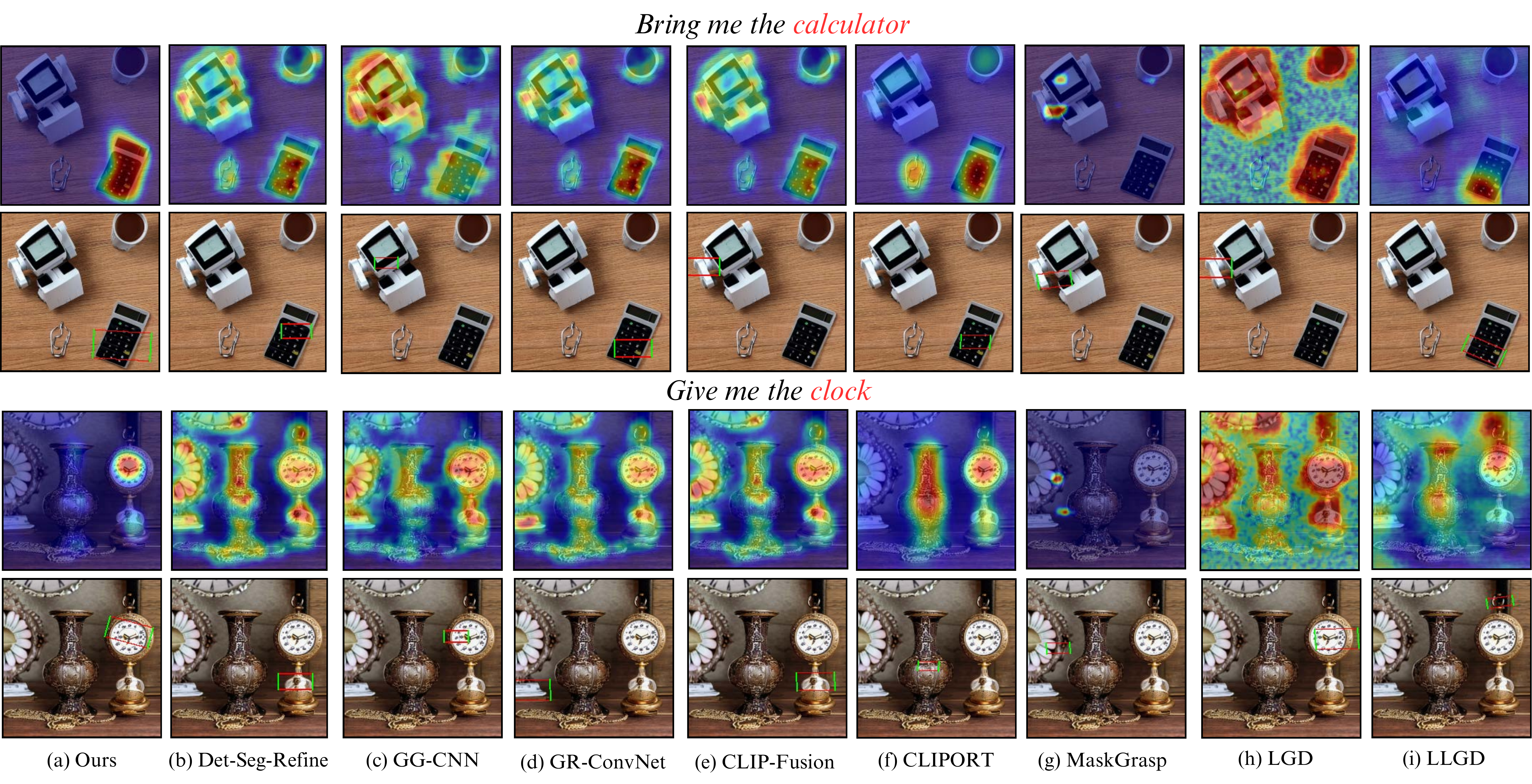}
 \vspace{-2ex}
	\caption{Visualization of language-driven grasp detection results of different methods.}
 \vspace{2ex}
\label{fig:vls_quantitive}
\end{figure*}
\subsection{Experimental Setup}
\textbf{Dataset.} 
To assess the generalization of all methods, we setup our experiments by training on the Grasp-Anything dataset~\cite{vuong2023grasp}. This dataset is created from large-scale foundation models which offers 1M images with textual descriptions. Following the approaches in~\cite{vuong2023grasp,zhou2022conditional}, we split the data into `Seen' and `Unseen' categories, designating $70\%$ of the categories as `Seen' and the remaining $30\%$ as `Unseen'. We also employ the harmonic mean (`H') metric to assess overall success rates~\cite{zhou2022conditional}.

\textbf{Evaluation Metrics.} Our primary evaluation metric is the success rate, defined similarly to~\cite{kumra2020antipodal}. A grasp is considered successful if the Intersection over Union (IoU) score between the predicted grasp and the ground truth exceeds $25\%$, and the offset angle is less than 30 degrees. During training, the text encoder is kept frozen, while the pre-trained vision backbone is fine-tuned using our dataset. We also measure the inference time of all methods using the same NVIDIA RTX 4080 GPU.

\textbf{Baselines.} We compare our method (GraspMamba) with GR-CNN~\cite{kumra2020antipodal}, Det-Seg-Refine~\cite{ainetter2021end}, GG-CNN~\cite{morrison2018closing}, CLIP-Fusion~\cite{xu2023joint}, MaskGrasp~\cite{vanvo2024languagedrivengraspdetectionmaskguided}, LGD~\cite{vuong2024language}, LLGD~\cite{nghia2024fastgrasp} and CLIPORT~\cite{shridhar2022cliport}, utilizing a pretrained CLIP~\cite{radford2021learning} model for text embedding.


\begin{table}[h]
\caption{\label{table:graspmamba_results}Language-driven grasp detection results.}
\centering
\renewcommand
\tabcolsep{4.5pt}
\vskip 0.1 in
\resizebox{\linewidth}{!}{
\begin{tabular}{@{}lcccccccc@{}}
\toprule
Baseline & Seen & UnSeen & H&Inference time\cr 
\midrule
Det-Seg-Refine~\cite{ainetter2021end} + CLIP~\cite{radford2021learning} &0.30 &0.15 &0.20&0.200s \\
GG-CNN~\cite{morrison2018closing} + CLIP~\cite{radford2021learning} &0.12 &0.08 &0.10&0.040s \\
GR-ConvNet~\cite{kumra2020antipodal} + CLIP~\cite{radford2021learning} &0.37 &0.18 &0.24&\textbf{0.022s}\\
CLIP-Fusion~\cite{xu2023joint} &0.40 &0.29 &0.33&0.157s\\
CLIPORT~\cite{shridhar2022cliport} &0.36 &0.26 &0.29&0.131s\\
LGD~\cite{vuong2024language} &0.48 &0.42 &0.45&22.00s\\
MaskGrasp~\cite{vanvo2024languagedrivengraspdetectionmaskguided} &0.50 &0.46 &0.45&0.116s\\
LLDG~\cite{nghia2024fastgrasp} &0.53 &0.39 &0.46&0.264s\\
\midrule
GraspMamba + BERT~\cite{devlin2018bert} (ours)& 0.60 &0.39 &0.46&0.032s\\
GraspMamba + CLIP~\cite{radford2021learning} (ours)& \textbf{0.63} &\textbf{0.41} &\textbf{0.52}&0.030s\\
\bottomrule
\end{tabular}}
\end{table}
\subsection{Language-driven Grasp Detection Results}
\textbf{Quantitative Results.} 
Table~\ref{table:graspmamba_results} summarises the results of all methods. From this table, we can see that our GraspMamba significantly outperforms other grasp detection techniques on the Grasp-Anything dataset. Our method consistently achieves better results than other baseline approaches in both the `Seen' and `Unseen' scenarios. Notably, in the `Seen' setup, our method shows substantial improvement, surpassing the second-best, LLGD~\cite{nghia2024fastgrasp} by a clear margin. Additionally, our inference time remains competitive, thanks to its simple architecture. Unlike diffusion-based methods~\cite{nghia2024fastgrasp, vuong2024language}, our GraspMamba demonstrates a good balance between accuracy and inference speed.

\textbf{Qualitative Results.} 
Fig.~\ref{fig:vls_quantitive} shows the quantitative evaluation of our method and other baselines. This figure shows that our method produces semantically plausible results, particularly in cluttered scenes with occlusions. The attention maps of our method also show an accurate connection between the text prompt and the visual region.

\subsection{Hierarchical Feature Fusion Analysis} 
\begin{figure}[h]
\centering
\includegraphics[width=0.99\linewidth]{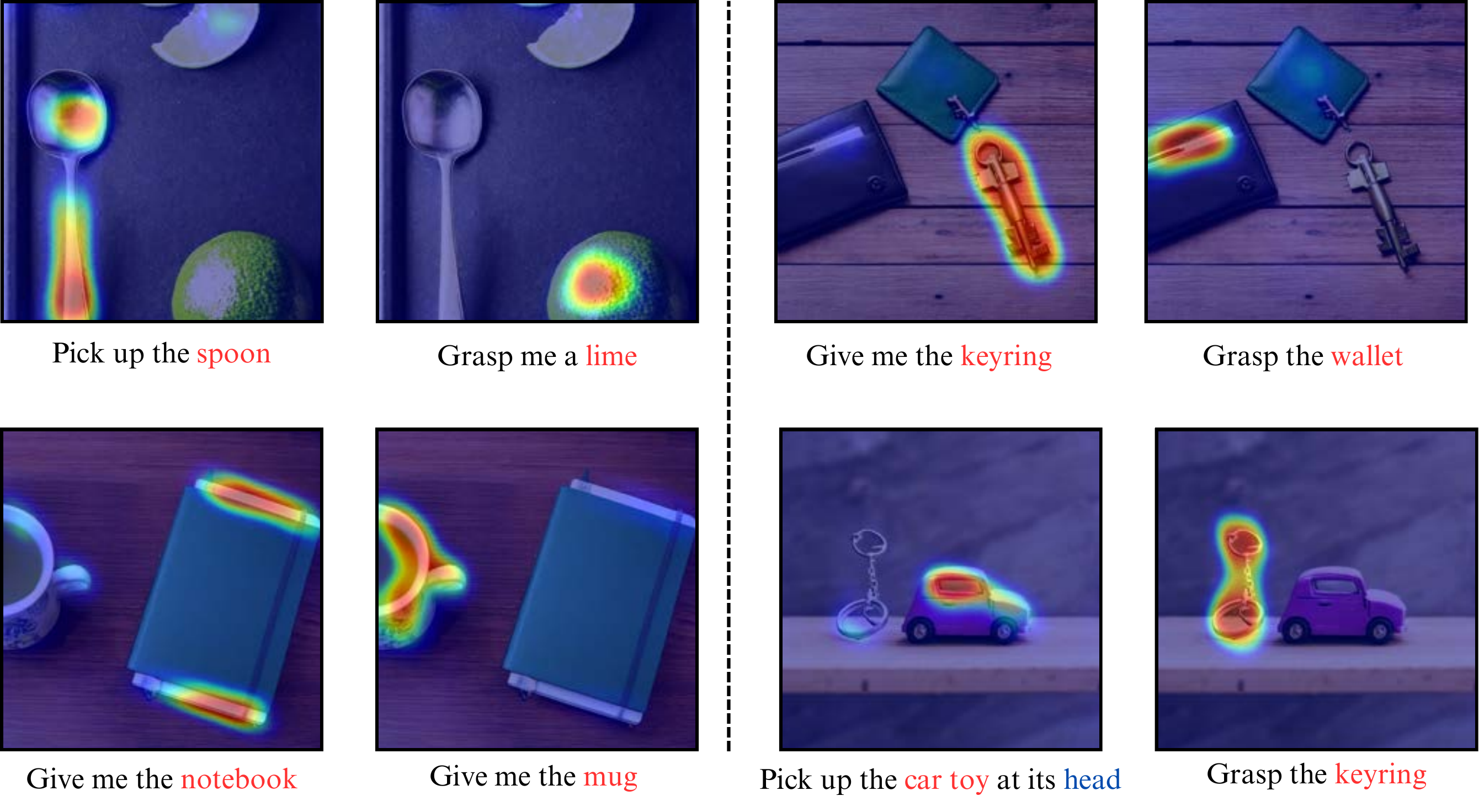}
\vspace{0.05ex}
\caption{The visualization of feature fusion result when different text inputs are used.}
\label{fig:vls_atten_fusion}
\end{figure}

To evaluate the performance of our hierarchical feature fusion block, we experiment with and without using the feature fusion block in our GraspMamba achitecture. Table~\ref{table:loss} summarises the results. We can see that the hierarchical feature fusion block has a positive impact on the grasp detection performance. Additionally, we visualize the attention maps generated by different methods using the text command. As shown in the heatmap row of Fig.~\ref{fig:vls_quantitive}, our approach effectively concentrates attention on the target object with minimal distraction from the surrounding area, distinguishing it from other methods. Our feature fusion technique successfully directs the model's focus toward essential regions, enabling the extraction of richer contextual information and enhancing grasp accuracy. While other methods tend to be less precise, with more background interference and a broader focus area. In addition, Fig.~\ref{fig:vls_atten_fusion} indicates that our method's effectiveness in aligning visual features with textual inputs. Overall, our method is able to locate and understand the referenced object based on textual instructions under the ambiguity and variability in different textual instructions while maintaining consistent visual alignment.

\begin{table}[t]
\caption{\label{table:loss} Feature Fusion Analysis.}
\vspace{-2ex}
\centering
\renewcommand
\tabcolsep{4.5pt}
\hspace{1ex}
\vskip 0.1 in

\begin{tabular}{@{}lcccccc@{}}
\toprule
Baseline & Seen & UnSeen & H\cr 
\midrule
GraspMamba without feature fusion & 0.582 &0.372 &0.477\\
GraspMamba with feature fusion& \textbf{0.630} &\textbf{0.411} &\textbf{0.521} \\
\bottomrule
\end{tabular}
\end{table}

\subsection{Ablation Study}

\begin{figure}[h]
\centering
\includegraphics[width=0.99\linewidth]{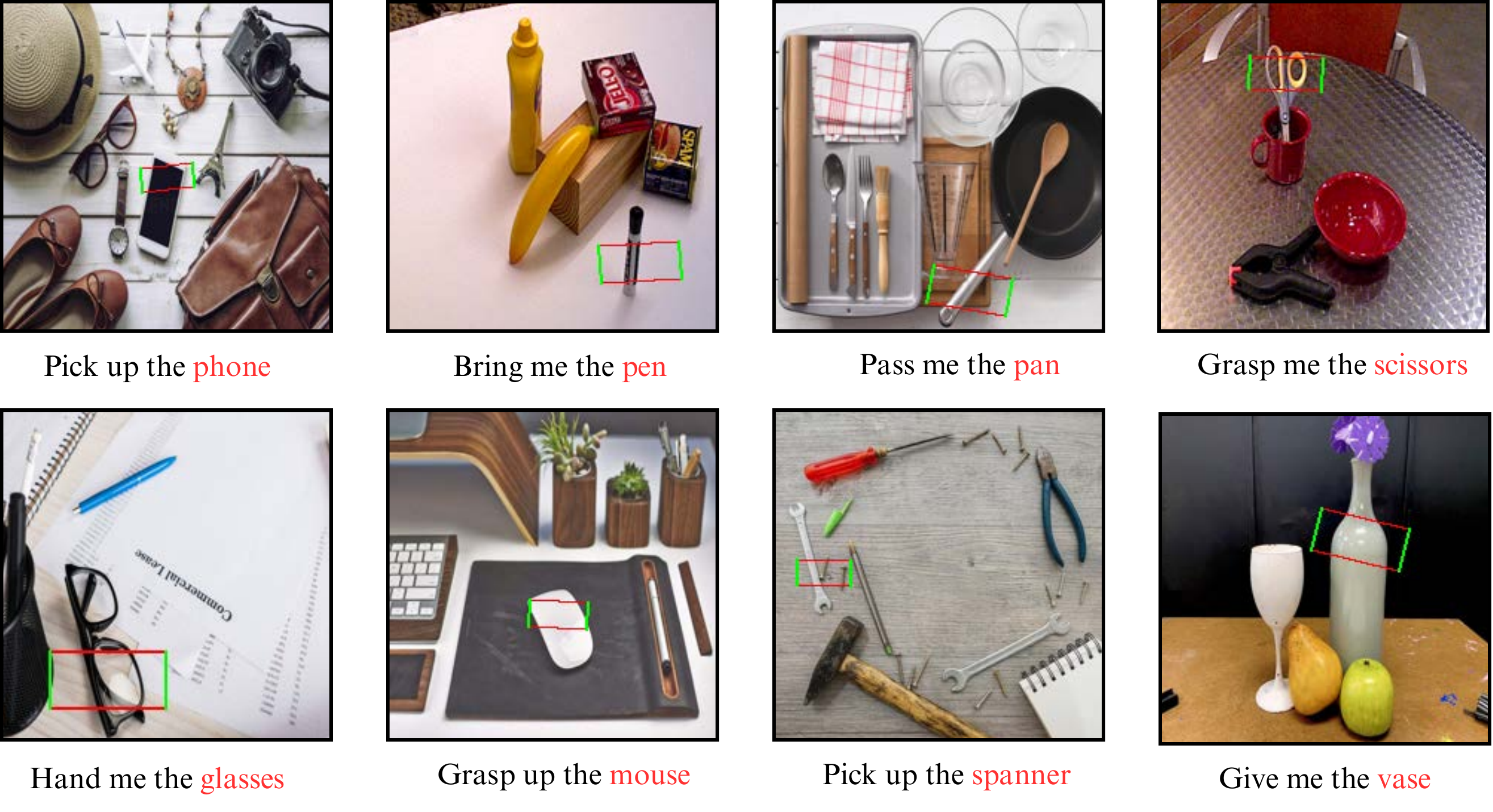}
\vspace{-2ex}
\caption{In the wild detection results. The images are from YCB-Video~\cite{fang2020graspnet} dataset and the internet.}
\vspace{2ex}
\label{fig:inthewild}
\end{figure}

\textbf{In the Wild Detection.} Fig.~\ref{fig:inthewild} presents in the wild visualization results by our method, which is exclusively trained on the Grasp-Anything dataset. These examples, applied to various random internet images and images from other datasets, illustrate our model's strong ability to generalize to real-world scenarios, even though it was trained entirely on synthetic data from Grasp-Anything dataset. This generalization is critical for deploying grasp detection systems in real-world applications, where the complexity and variability of environments often differ from training datasets.

\begin{figure}[h]
\centering
\includegraphics[width=0.99\linewidth]{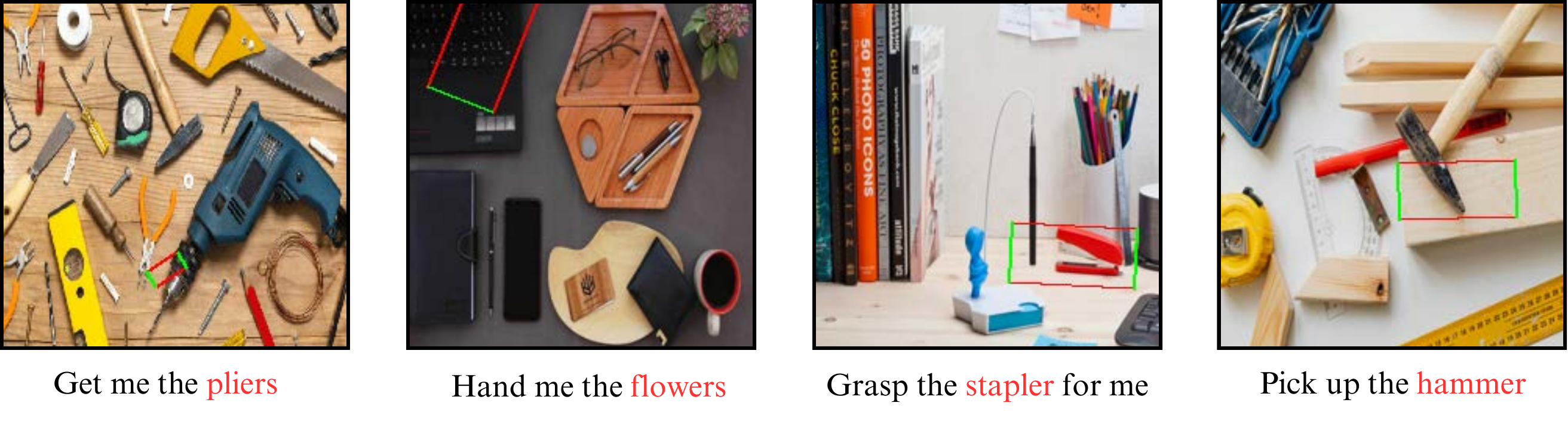}
\vspace{-2ex}
\caption{Failure cases of our method.}
\label{fig:failures}
\end{figure}

\textbf{Failure Cases.} Despite successfully improving the alignment of textual and visual features, our method still produces incorrect grasping poses in some cases. The wide variety of objects and grasping prompts presents a significant challenge, as the network struggles to capture the full range of real-life scenarios. Fig.~\ref{fig:failures} illustrates several failure cases where our GraspMamba makes incorrect predictions. In some instances, the text prompt is aligned with the correct object, but the grasping pose remains inaccurate. This can be attributed to the presence of multiple similar objects or shapes that are difficult to differentiate and to text prompts that lack sufficient detail for precise predictions.

\subsection{Robotic Validation}

In Fig.~\ref{fig: robot demonstration}, we present our robotic evaluation using a Kinova Gen3 7-DoF robot. The grasp detection, along with other techniques listed in Table~\ref{table: real-robot-exp}, is tested using depth images from an Intel RealSense D410 depth camera. Our method predicts 4-DoF grasping poses, which are converted to 6-DoF poses, assuming objects are placed on flat surfaces. Trajectory optimization, as outlined in~\cite{beck2023singularity,vu2023machine}, directs the robot toward the desired poses. The inference and control processes run on an Intel Core i7 12700K processor and an NVIDIA RTX 4080S Ti graphics card. We evaluate performance in both single-object and cluttered environments with a variety of real-world objects, repeating each test $25$ times for consistency across all methods. As highlighted in Table~\ref{table: real-robot-exp}, our approach, utilizing Mamba architecture, consistently outperforms other baselines. Notably, despite being trained exclusively on Grasp-Anything, a synthetic dataset generated using foundational models, it demonstrates strong performance on real-world objects.

%% file: 5_conclusions.tex
\section{Discussion}\label{Sec:con}
\textbf{Limitation.} While our method demonstrates notable results in terms of inference time and accuracy, it still struggles with predicting correct grasping poses in scenes involving complex real-world images. Although our approach shows promise in identifying the attention area on the target object, faulty grasping poses often arise due to ambiguities regarding the grasping part, as shown in Fig.~\ref{fig:failures}. Our experiments show that when grasp instruction sentences contain rare or complex nouns, ambiguity in parsing or interpreting the text prompts often arises. This leads to incorrect grasp predictions. For example, in the instruction ``grasp me the stapler" the model struggles to distinguish the stapler from a nearby pencil or fails to locate a flower. Therefore, providing clear and precise instruction prompts is crucial for enabling the robot to understand and execute accurate grasping actions.

\begin{figure}[t]
\centering
\def\svgwidth{1\columnwidth}
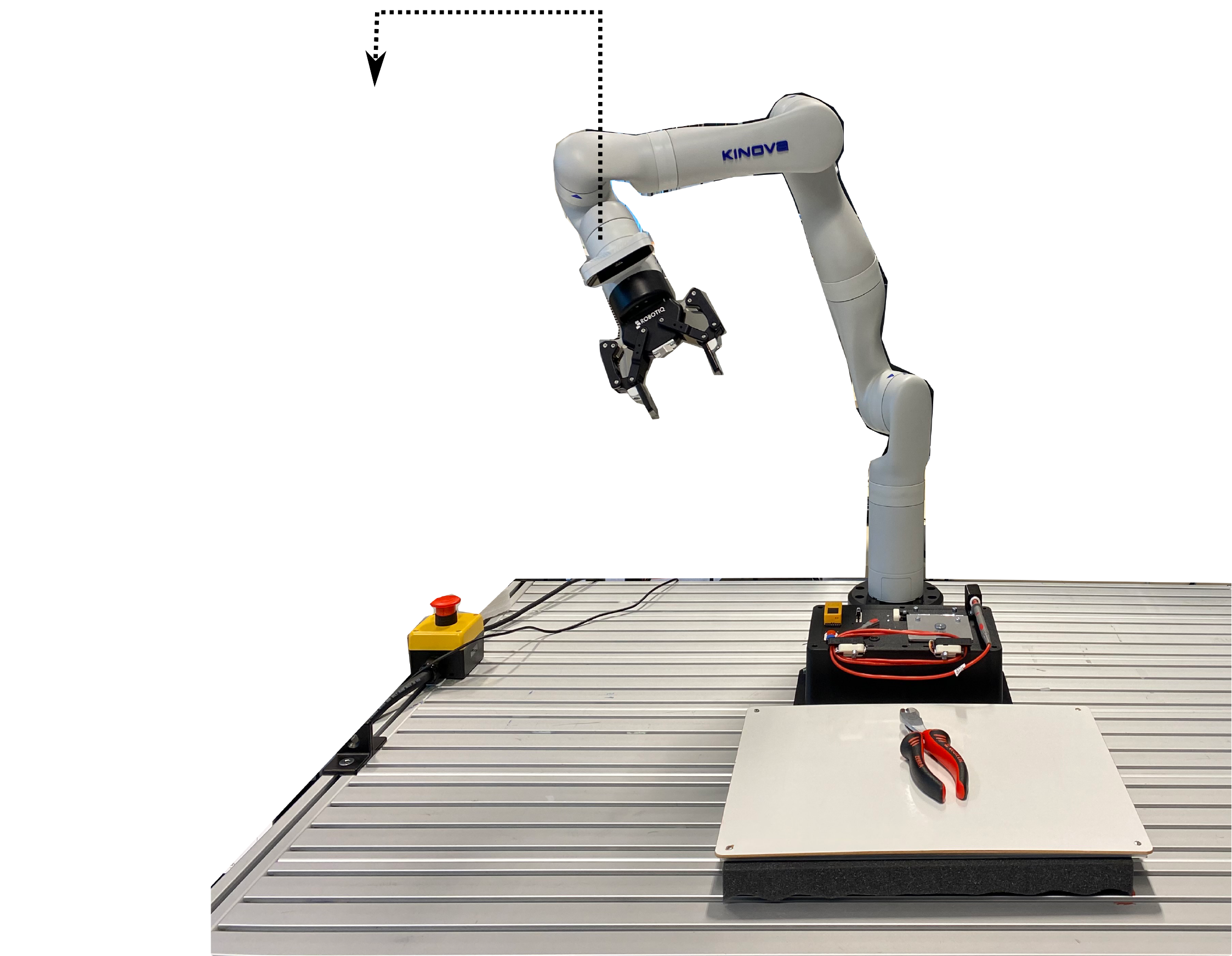
\vspace{-2mm}
\caption{The robotic experiment setup.}
\label{fig: robot demonstration}
\end{figure}

\begin{table}[t]
    \centering
    \caption{\label{table: real-robot-exp} Robotic language-driven grasp detection results}
    \vspace{2ex}
    \renewcommand
\tabcolsep{4pt}
\hspace{1ex}
    \begin{tabular}{@{}rcc@{}}
\toprule
Baseline & Single &  Cluttered\cr 
\midrule
Det-Seg-Refine~\cite{ainetter2021end} + CLIP~\cite{radford2021learning} &0.30  & 0.23\\
GG-CNN~\cite{morrison2018closing} + CLIP~\cite{radford2021learning} &0.10  & 0.07 \\
GR-ConvNet~\cite{kumra2020antipodal} + CLIP~\cite{radford2021learning}  &0.33  & 0.30\\
CLIP-Fusion~\cite{xu2023joint} & 0.40 & 0.40 \\
CLIPORT~\cite{shridhar2022cliport} &0.27 & 0.30 \\
LLGD~\cite{nghia2024fastgrasp}  &0.43 & 0.42 \\
MaskGrasp~\cite{vanvo2024languagedrivengraspdetectionmaskguided}  &0.43 & 0.42 \\
\midrule
GraspMamba (ours) &  \textbf{0.54} & \textbf{0.52} \\
\bottomrule
\end{tabular}
\end{table}

\textbf{Future work.} While our study proposes a new method that integrates textual and visual features using the Mamba architecture for grasping pose detection. We see several prospects for improvement in future work. First, we aim to extend our method to handle tasks in 3D space, including 3D point clouds and RGB-D images, to overcome the limitations of depth information in robotic applications. Additionally, bridging the gap between the semantic concepts in text prompts and input images can enhance speed, efficiency, and hardware optimization, especially for processing long sequences. For instance, enabling the robot to comprehend and analyze complex instructions from humans and make accurate decisions quickly, without relying on high-powered, energy-inefficient hardware, is a key goal. These approaches offer significant potential for advancing the capabilities of language-driven robotic grasping systems.

\section{Conclusion}
We introduce a new vision-language model based on the Mamba vision architecture for the language-driven grasp detection task. Our approach employs hierarchical feature fusion of text and image inputs, effectively integrating visual and textual information to improve both grasping accuracy and inference speed. By focusing on key regions highlighted by text guidance prompts, our method achieves high precision. Extensive experiments demonstrate that our approach significantly outperforms existing baselines in vision-based benchmarks and real-world robotic grasping tests. Our code and model will be released.

%% file: robotic-experiment-mamba.pdf_tex
\begingroup%
  \makeatletter%
  \providecommand\color[2][]{%
    \errmessage{(Inkscape) Color is used for the text in Inkscape, but the package 'color.sty' is not loaded}%
    \renewcommand\color[2][]{}%
  }%
  \providecommand\transparent[1]{%
    \errmessage{(Inkscape) Transparency is used (non-zero) for the text in Inkscape, but the package 'transparent.sty' is not loaded}%
    \renewcommand\transparent[1]{}%
  }%
  \providecommand\rotatebox[2]{#2}%
  \newcommand*\fsize{\dimexpr\f@size pt\relax}%
  \newcommand*\lineheight[1]{\fontsize{\fsize}{#1\fsize}\selectfont}%
  \ifx\svgwidth\undefined%
    \setlength{\unitlength}{1078.10074934bp}%
    \ifx\svgscale\undefined%
      \relax%
    \else%
      \setlength{\unitlength}{\unitlength * \real{\svgscale}}%
    \fi%
  \else%
    \setlength{\unitlength}{\svgwidth}%
  \fi%
  \global\let\svgwidth\undefined%
  \global\let\svgscale\undefined%
  \makeatother%
  \begin{picture}(1,0.77634441)%
    \lineheight{1}%
    \setlength\tabcolsep{0pt}%
    \put(0,0){\includegraphics[width=\unitlength,page=1]{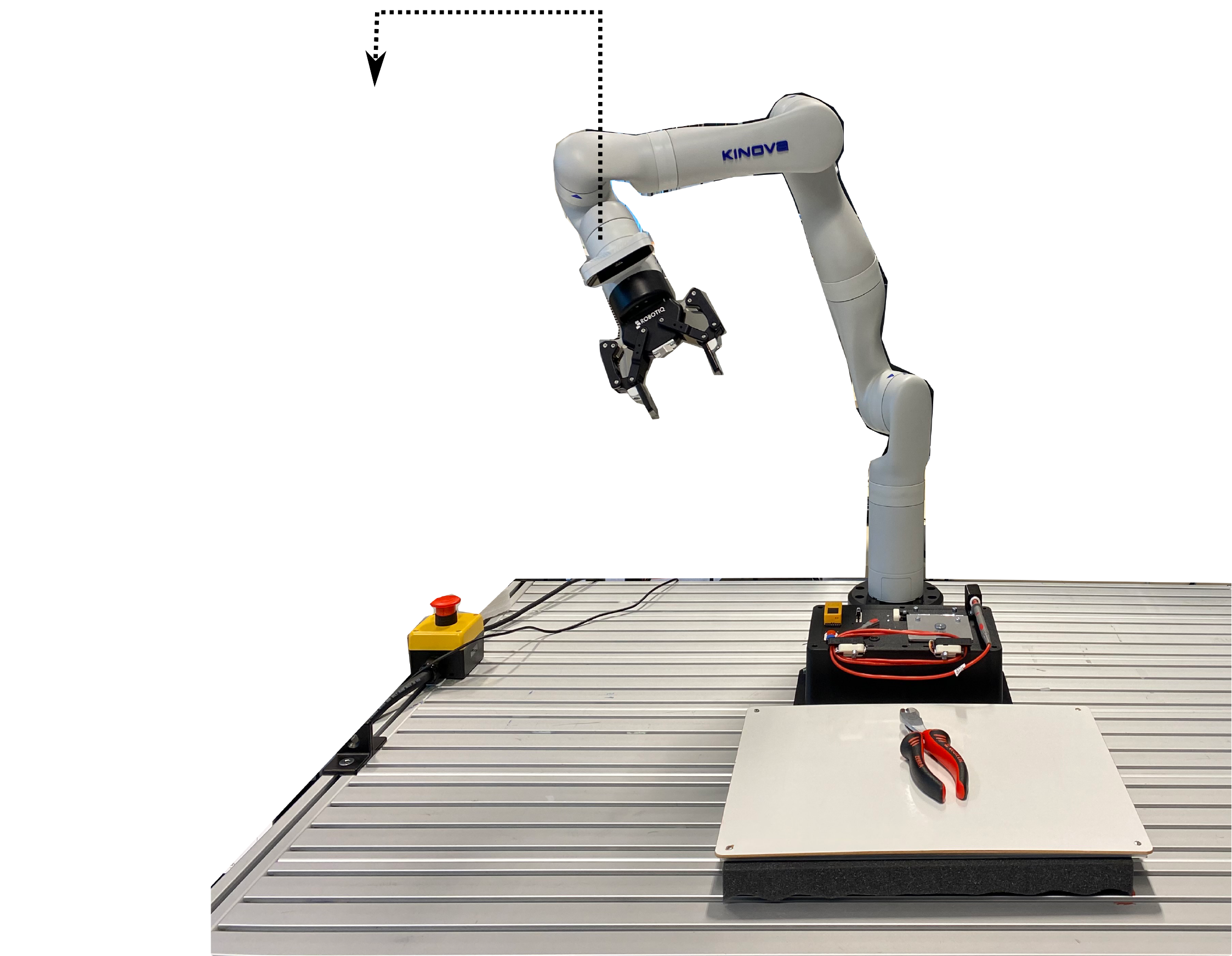}}%
    \put(0.49959526,0.74182284){\color[rgb]{0,0,0}\makebox(0,0)[lt]{\lineheight{1.25}\smash{\begin{tabular}[t]{l}Realsense D410\end{tabular}}}}%
    \put(0.43793179,0.3336237){\color[rgb]{0,0,0}\makebox(0,0)[lt]{\lineheight{1.25}\smash{\begin{tabular}[t]{l}Robotiq 2F-85\end{tabular}}}}%
    \put(0.10695573,0.67137718){\color[rgb]{0,0,0}\makebox(0,0)[lt]{\lineheight{1.25}\smash{\begin{tabular}[t]{l}\scalebox{0.9}{GraspMamba}\end{tabular}}}}%
    \put(0,0){\includegraphics[width=\unitlength,page=2]{robotic-experiment-mamba.pdf}}%
    \put(0.15079903,0.35074012){\color[rgb]{0,0,0}\makebox(0,0)[lt]{\lineheight{1.25}\smash{\begin{tabular}[t]{l}\scalebox{0.9}{ROS 2}\end{tabular}}}}%
    \put(0.0817604,0.29673434){\color[rgb]{0,0,0}\makebox(0,0)[lt]{\lineheight{1.25}\smash{\begin{tabular}[t]{l}\scalebox{0.9}{Grapsing pose}\\\end{tabular}}}}%
    \put(0.12788537,0.18877209){\color[rgb]{0,0,0}\makebox(0,0)[lt]{\lineheight{1.25}\smash{\begin{tabular}[t]{l}\scalebox{0.9}{\textcolor{white}{Trajectory}}\\\end{tabular}}}}%
    \put(0.10804481,0.15816606){\color[rgb]{0,0,0}\makebox(0,0)[lt]{\lineheight{1.25}\smash{\begin{tabular}[t]{l}\scalebox{0.9}{\textcolor{white}{optimization}}\\\end{tabular}}}}%
    \put(0.13218615,0.25858587){\color[rgb]{0,0,0}\makebox(0,0)[lt]{\lineheight{1.25}\smash{\begin{tabular}[t]{l}\scalebox{0.9}{generation}\\\end{tabular}}}}%
    \put(0,0){\includegraphics[width=\unitlength,page=3]{robotic-experiment-mamba.pdf}}%
    \put(0.07671737,0.03384816){\color[rgb]{0,0,0}\makebox(0,0)[lt]{\lineheight{1.25}\smash{\begin{tabular}[t]{l}\scalebox{0.8}{\textcolor{green}{Torque controller}}\end{tabular}}}}%
    \put(0,0){\includegraphics[width=\unitlength,page=4]{robotic-experiment-mamba.pdf}}%
    \put(-0.00267667,0.7552027){\color[rgb]{0,0,0}\makebox(0,0)[lt]{\lineheight{1.25}\smash{\begin{tabular}[t]{l}\scalebox{0.9}{``Grasp the pliers"}\end{tabular}}}}%
    \put(0,0){\includegraphics[width=\unitlength,page=5]{robotic-experiment-mamba.pdf}}%
  \end{picture}%
\endgroup%

%% file: MAIN.bbl
\begin{thebibliography}{10}
\providecommand{\url}[1]{#1}
\csname url@rmstyle\endcsname
\providecommand{\newblock}{\relax}
\providecommand{\bibinfo}[2]{#2}
\providecommand\BIBentrySTDinterwordspacing{\spaceskip=0pt\relax}
\providecommand\BIBentryALTinterwordstretchfactor{4}
\providecommand\BIBentryALTinterwordspacing{\spaceskip=\fontdimen2\font plus
\BIBentryALTinterwordstretchfactor\fontdimen3\font minus \fontdimen4\font\relax}
\providecommand\BIBforeignlanguage[2]{{%
\expandafter\ifx\csname l@#1\endcsname\relax
\typeout{** WARNING: IEEEtran.bst: No hyphenation pattern has been}%
\typeout{** loaded for the language `#1'. Using the pattern for}%
\typeout{** the default language instead.}%
\else
\language=\csname l@#1\endcsname
\fi
#2}}

\bibitem{sundermeyer2021contact}
M.~Sundermeyer, A.~Mousavian, R.~Triebel, and D.~Fox, ``Contact-graspnet: Efficient 6-dof grasp generation in cluttered scenes,'' in \emph{ICRA}, 2021.

\bibitem{wen2022catgrasplearning}
B.~Wen, W.~Lian, K.~Bekris, and S.~Schaal, ``Catgrasp: Learning category-level task-relevant grasping in clutter from simulation,'' in \emph{ICRA}, 2022.

\bibitem{nguyen2016preparatory}
A.~Nguyen, D.~Kanoulas, D.~G. Caldwell, and N.~G. Tsagarakis, ``Preparatory object reorientation for task-oriented grasping,'' in \emph{IROS}, 2016.

\bibitem{ainetter2022endtoend}
S.~Ainetter and F.~Fraundorfer, ``End-to-end trainable deep neural network for robotic grasp detection and semantic segmentation from rgb,'' in \emph{ICRA}, 2022.

\bibitem{redmon2015realtime}
J.~Redmon and A.~Angelova, ``Real-time grasp detection using convolutional neural networks,'' in \emph{ICRA}, 2015.

\bibitem{gilles2023metagraspnetv2}
M.~Gilles \emph{et~al.}, ``Metagraspnetv2: All-in-one dataset enabling fast and reliable robotic bin picking via object relationship reasoning and dexterous grasping,'' \emph{TASE}, 2023.

\bibitem{schirmer2024towards}
F.~Schirmer, P.~Kranz, B.~Bhat, C.~G. Rose, J.~Schmitt, and T.~Kaupp, ``Towards a path planning and communication framework for seamless human-robot assembly,'' in \emph{HRI}, 2024.

\bibitem{vuong2023grasp}
A.~D. Vuong \emph{et~al.}, ``Grasp-anything: Large-scale grasp dataset from foundation models,'' in \emph{ICRA}, 2024.

\bibitem{yuan2023m2t2}
W.~Yuan, A.~Murali, A.~Mousavian, and D.~Fox, ``M2t2: Multi-task masked transformer for object-centric pick and place,'' in \emph{CoRL}, 2023.

\bibitem{vuong2024language}
A.~D. Vuong, M.~N. Vu, B.~Huang, N.~Nguyen, H.~Le, T.~Vo, and A.~Nguyen, ``Language-driven grasp detection,'' in \emph{CVPR}, 2024.

\bibitem{yuan2024robopoint}
W.~Yuan, J.~Duan, V.~Blukis, W.~Pumacay, R.~Krishna, A.~Murali, A.~Mousavian, and D.~Fox, ``Robopoint: A vision-language model for spatial affordance prediction for robotics,'' \emph{arXiv}, 2024.

\bibitem{o2024open}
A.~O’Neill \emph{et~al.}, ``Open x-embodiment: Robotic learning datasets and rt-x models: Open x-embodiment collaboration 0,'' in \emph{ICRA}, 2024.

\bibitem{vanvo2024languagedrivengraspdetectionmaskguided}
T.~V. Vo, M.~N. Vu, B.~Huang, A.~Vuong, N.~Le, T.~Vo, and A.~Nguyen, ``Language-driven grasp detection with mask-guided attention,'' in \emph{IROS}, 2024.

\bibitem{nghia2024fastgrasp}
N.~Nguyen, M.~N. Vu, B.~Huang, A.~Vuong, N.~Le, T.~Vo, and A.~Nguyen, ``Lightweight language-driven grasp detection using conditional consistency model,'' in \emph{IROS}, 2024.

\bibitem{ahn2022icanisay}
M.~Ahn, A.~Brohan, N.~Brown, Y.~Chebotar, O.~Cortes, B.~David, C.~Finn, C.~Fu, K.~Gopalakrishnan, K.~Hausman, A.~Herzog, D.~Ho, J.~Hsu, J.~Ibarz, B.~Ichter, A.~Irpan, E.~Jang, R.~J. Ruano, K.~Jeffrey, S.~Jesmonth, N.~J. Joshi, R.~Julian, D.~Kalashnikov, Y.~Kuang, K.-H. Lee, S.~Levine, Y.~Lu, L.~Luu, C.~Parada, P.~Pastor, J.~Quiambao, K.~Rao, J.~Rettinghouse, D.~Reyes, P.~Sermanet, N.~Sievers, C.~Tan, A.~Toshev, V.~Vanhoucke, F.~Xia, T.~Xiao, P.~Xu, S.~Xu, M.~Yan, and A.~Zeng, ``Do as i can, not as i say: Grounding language in robotic affordances,'' \emph{arXiv}, 2022.

\bibitem{driess2023palm}
D.~Driess \emph{et~al.}, ``Palm-e: An embodied multimodal language model,'' \emph{arXiv}, 2023.

\bibitem{chatgpt}
OpenAI, ``{Introducing ChatGPT},'' Software, accessed: February 6th 2023.

\bibitem{shridhar2022cliport}
M.~Shridhar \emph{et~al.}, ``Cliport: What and where pathways for robotic manipulation,'' in \emph{CoRL}, 2022.

\bibitem{xu2023jointmodelingvision}
K.~Xu, S.~Zhao, Z.~Zhou, Z.~Li, H.~Pi, Y.~Zhu, Y.~Wang, and R.~Xiong, ``A joint modeling of vision-language-action for target-oriented grasping in clutter,'' \emph{arXiv}, 2023.

\bibitem{mirjalili2023langraspusinglargelanguage}
R.~Mirjalili, M.~Krawez, S.~Silenzi, Y.~Blei, and W.~Burgard, ``Lan-grasp: Using large language models for semantic object grasping,'' \emph{arXiv}, 2023.

\bibitem{tang2023graspgpt}
C.~Tang, D.~Huang, W.~Ge, W.~Liu, and H.~Zhang, ``Graspgpt: Leveraging semantic knowledge from a large language model for task-oriented grasping,'' \emph{arXiv}, 2023.

\bibitem{gu2024mamba}
A.~Gu and T.~Dao, ``Mamba: Linear-time sequence modeling with selective state spaces,'' \emph{arXiv}, 2024.

\bibitem{han2024mamba3d}
X.~Han, Y.~Tang, Z.~Wang, and X.~Li, ``Mamba3d: Enhancing local features for 3d point cloud analysis via state space model,'' \emph{arXiv}, 2024.

\bibitem{liu2024robomamba}
J.~Liu, M.~Liu, Z.~Wang, L.~Lee, K.~Zhou, P.~An, S.~Yang, R.~Zhang, Y.~Guo, and S.~Zhang, ``Robomamba: Multimodal state space model for efficient robot reasoning and manipulation,'' \emph{arXiv}, 2024.

\bibitem{lieber2024jamba}
O.~Lieber, B.~Lenz, H.~Bata, G.~Cohen, J.~Osin, I.~Dalmedigos, E.~Safahi, S.~Meirom, Y.~Belinkov, S.~Shalev-Shwartz, O.~Abend, R.~Alon, T.~Asida, A.~Bergman, R.~Glozman, M.~Gokhman, A.~Manevich, N.~Ratner, N.~Rozen, E.~Shwartz, M.~Zusman, and Y.~Shoham, ``Jamba: A hybrid transformer-mamba language model,'' \emph{arXiv}, 2024.

\bibitem{chen2024resvmamba}
C.-S. Chen, G.-Y. Chen, D.~Zhou, D.~Jiang, and D.-S. Chen, ``Res-vmamba: Fine-grained food category visual classification using selective state space models with deep residual learning,'' \emph{arXiv}, 2024.

\bibitem{chen2024rsmambaremotesensingimage}
K.~Chen, B.~Chen, C.~Liu, W.~Li, Z.~Zou, and Z.~Shi, ``Rsmamba: Remote sensing image classification with state space model,'' \emph{arXiv}, 2024.

\bibitem{wang2024s2mambaspatialspectralstatespace}
G.~Wang, X.~Zhang, Z.~Peng, T.~Zhang, and L.~Jiao, ``S$^2$mamba: A spatial-spectral state space model for hyperspectral image classification,'' \emph{arXiv}, 2024.

\bibitem{ma2024rs3mambavisualstatespace}
X.~Ma, X.~Zhang, and M.-O. Pun, ``Rs3mamba: Visual state space model for remote sensing images semantic segmentation,'' \emph{arXiv}, 2024.

\bibitem{zhu2024samba}
Q.~Zhu, Y.~Cai, Y.~Fang, Y.~Yang, C.~Chen, L.~Fan, and A.~Nguyen, ``Samba: Semantic segmentation of remotely sensed images with state space model,'' \emph{arXiv}, 2024.

\bibitem{wan2024sigmasiamesemambanetwork}
Z.~Wan, Y.~Wang, S.~Yong, P.~Zhang, S.~Stepputtis, K.~Sycara, and Y.~Xie, ``Sigma: Siamese mamba network for multi-modal semantic segmentation,'' \emph{arXiv}, 2024.

\bibitem{liang2024pointmamba}
D.~Liang, X.~Zhou, W.~Xu, X.~Zhu, Z.~Zou, X.~Ye, X.~Tan, and X.~Bai, ``Pointmamba: A simple state space model for point cloud analysis,'' \emph{arXiv}, 2024.

\bibitem{wang2024pointramba}
Z.~Wang, Z.~Chen, Y.~Wu, Z.~Zhao, L.~Zhou, and D.~Xu, ``Pointramba: A hybrid transformer-mamba framework for point cloud analysis,'' \emph{arXiv}, 2024.

\bibitem{jia2024mail}
X.~Jia, Q.~Wang, A.~Donat, B.~Xing, G.~Li, H.~Zhou, O.~Celik, D.~Blessing, R.~Lioutikov, and G.~Neumann, ``Mail: Improving imitation learning with mamba,'' \emph{arXiv}, 2024.

\bibitem{fang2019scene}
K.~Fang, A.~Toshev, L.~Fei-Fei, and S.~Savarese, ``Scene memory transformer for embodied agents in long-horizon tasks,'' in \emph{CVPR}, 2019.

\bibitem{domae2014fast}
Y.~Domae \emph{et~al.}, ``Fast graspability evaluation on single depth maps for bin picking with general grippers,'' in \emph{ICRA}, 2014.

\bibitem{bohg2014data}
J.~Bohg, A.~Morales, T.~Asfour, and D.~Kragic, ``Data-driven grasp synthesis—a survey,'' \emph{IEEE Transactions on Robotics}, 2014.

\bibitem{zhou2018fullyconvolutional}
X.~Zhou, X.~Lan, H.~Zhang, Z.~Tian, Y.~Zhang, and N.~Zheng, ``Fully convolutional grasp detection network with oriented anchor box,'' in \emph{IROS}, 2018.

\bibitem{yu2023egnet}
S.~Yu, D.-H. Zhai, and Y.~Xia, ``Egnet: Efficient robotic grasp detection network,'' \emph{IEEE Transactions on Industrial Electronics}, 2023.

\bibitem{guo2017ahybrid}
D.~Guo, F.~Sun, H.~Liu, T.~Kong, B.~Fang, and N.~Xi, ``A hybrid deep architecture for robotic grasp detection,'' in \emph{ICRA}, 2017.

\bibitem{yu2022seresunet}
S.~Yu, D.-H. Zhai, Y.~Xia, H.~Wu, and J.~Liao, ``Se-resunet: A novel robotic grasp detection method,'' \emph{IEEE Robotics and Automation Letters}, 2022.

\bibitem{tong2024anovelrgbd}
L.~Tong, K.~Song, H.~Tian, Y.~Man, Y.~Yan, and Q.~Meng, ``A novel rgb-d cross-background robot grasp detection dataset and background-adaptive grasping network,'' \emph{IEEE Transactions on Instrumentation and Measurement}, 2024.

\bibitem{wang2021graspness}
C.~Wang, H.-S. Fang, M.~Gou, H.~Fang, J.~Gao, and C.~Lu, ``Graspness discovery in clutters for fast and accurate grasp detection,'' in \emph{ICCV}, 2021.

\bibitem{ni2020pointnet}
P.~Ni, W.~Zhang, X.~Zhu, and Q.~Cao, ``Pointnet++ grasping: Learning an end-to-end spatial grasp generation algorithm from sparse point clouds,'' in \emph{ICRA}, 2020.

\bibitem{nguyen2023open}
T.~Nguyen, M.~N. Vu, A.~Vuong, D.~Nguyen, T.~Vo, N.~Le, and A.~Nguyen, ``Open-vocabulary affordance detection in 3d point clouds,'' in \emph{IROS}, 2023.

\bibitem{nguyen2024LGrasp6D}
T.~Nguyen, M.~N. Vu, B.~Huang, A.~Vuong, Q.~Vuong, N.~Le, T.~Vo, and A.~Nguyen, ``Language-driven 6-dof grasp detection using negative prompt guidance,'' in \emph{ECCV}, 2024.

\bibitem{lu2023vl}
Y.~Lu \emph{et~al.}, ``Vl-grasp: a 6-dof interactive grasp policy for language-oriented objects in cluttered indoor scenes,'' in \emph{IROS}, 2023.

\bibitem{sun2023language}
Q.~Sun, H.~Lin, Y.~Fu, Y.~Fu, and X.~Xue, ``Language guided robotic grasping with fine-grained instructions,'' in \emph{IROS}, 2023.

\bibitem{cheang2022learning}
C.~Cheang, H.~Lin, Y.~Fu, and X.~Xue, ``Learning 6-dof object poses to grasp category-level objects by language instructions,'' in \emph{ICRA}, 2022.

\bibitem{xu2023joint}
K.~Xu \emph{et~al.}, ``A joint modeling of vision-language-action for target-oriented grasping in clutter,'' \emph{arXiv}, 2023.

\bibitem{yang2022interactive}
Y.~Yang \emph{et~al.}, ``Interactive robotic grasping with attribute-guided disambiguation,'' \emph{arXiv}, 2022.

\bibitem{jin2024reasoning}
S.~Jin, J.~Xu, Y.~Lei, and L.~Zhang, ``Reasoning grasping via multimodal large language model,'' \emph{arXiv}, 2024.

\bibitem{huang2016instanceaware}
Y.~Huang, W.~Wang, and L.~Wang, ``Instance-aware image and sentence matching with selective multimodal lstm,'' in \emph{CVPR}, 2016.

\bibitem{huang2017learning}
Y.~Huang, Q.~Wu, and L.~Wang, ``Learning semantic concepts and order for image and sentence matching,'' in \emph{CVPR}, 2017.

\bibitem{xu2021crossmodal}
X.~Xu, Y.~Wang, Y.~He, Y.~Yang, A.~Hanjalic, and H.~T. Shen, ``Cross-modal hybrid feature fusion for image-sentence matching,'' \emph{ACM Trans. Multimedia Comput. Commun. Appl.}, 2021.

\bibitem{vaswani2017attention}
A.~Vaswani \emph{et~al.}, ``Attention is all you need,'' \emph{NeurIPS}, 2017.

\bibitem{zhu2024visionmamba}
L.~Zhu, B.~Liao, Q.~Zhang, X.~Wang, W.~Liu, and X.~Wang, ``Vision mamba: Efficient visual representation learning with bidirectional state space model,'' \emph{arXiv}, 2024.

\bibitem{yang2024plainmamba}
C.~Yang, Z.~Chen, M.~Espinosa, L.~Ericsson, Z.~Wang, J.~Liu, and E.~J. Crowley, ``Plainmamba: Improving non-hierarchical mamba in visual recognition,'' \emph{arXiv}, 2024.

\bibitem{pei2024efficientvmamba}
X.~Pei, T.~Huang, and C.~Xu, ``Efficientvmamba: Atrous selective scan for light weight visual mamba,'' \emph{arXiv}, 2024.

\bibitem{liu2024vmamba}
Y.~Liu, Y.~Tian, Y.~Zhao, H.~Yu, L.~Xie, Y.~Wang, Q.~Ye, and Y.~Liu, ``Vmamba: Visual state space model,'' \emph{arXiv}, 2024.

\bibitem{hatamizadeh2024mambavision}
A.~Hatamizadeh and J.~Kautz, ``Mambavision: A hybrid mamba-transformer vision backbone,'' \emph{arXiv}, 2024.

\bibitem{zhao2024cobra}
H.~Zhao, M.~Zhang, W.~Zhao, P.~Ding, S.~Huang, and D.~Wang, ``Cobra: Extending mamba to multi-modal large language model for efficient inference,'' \emph{arXiv}, 2024.

\bibitem{qiao2024vlmamba}
Y.~Qiao, Z.~Yu, L.~Guo, S.~Chen, Z.~Zhao, M.~Sun, Q.~Wu, and J.~Liu, ``Vl-mamba: Exploring state space models for multimodal learning,'' \emph{arXiv}, 2024.

\bibitem{xu2024survey}
R.~Xu, S.~Yang, Y.~Wang, B.~Du, and H.~Chen, ``A survey on vision mamba: Models, applications and challenges,'' \emph{arXiv preprint arXiv:2404.18861}, 2024.

\bibitem{liu2021swin}
Z.~Liu, Y.~Lin, Y.~Cao, H.~Hu, Y.~Wei, Z.~Zhang, S.~Lin, and B.~Guo, ``Swin transformer: Hierarchical vision transformer using shifted windows,'' in \emph{ICCV}, 2021.

\bibitem{zhong2023multi}
G.~Zhong, W.~Ding, L.~Chen, Y.~Wang, and Y.-F. Yu, ``Multi-scale attention generative adversarial network for medical image enhancement,'' \emph{TETCI}, 2023.

\bibitem{depierre2018jacquard}
A.~Depierre, E.~Dellandr{\'e}a, and L.~Chen, ``Jacquard: A large scale dataset for robotic grasp detection,'' in \emph{IROS}, 2018.

\bibitem{devlin2018bert}
J.~Devlin \emph{et~al.}, ``Bert: Pre-training of deep bidirectional transformers for language understanding,'' \emph{arXiv}, 2018.

\bibitem{radford2021learning}
A.~Radford \emph{et~al.}, ``Learning transferable visual models from natural language supervision,'' in \emph{ICML}, 2021.

\bibitem{li2024mambadfusel}
Z.~Li, H.~Pan, K.~Zhang, Y.~Wang, and F.~Yu, ``Mambadfuse: A mamba-based dual-phase model for multi-modality image fusion,'' \emph{arXiv}, 2024.

\bibitem{li2024coupledmamba}
W.~Li, H.~Zhou, J.~Yu, Z.~Song, and W.~Yang, ``Coupled mamba: Enhanced multi-modal fusion with coupled state space model,'' \emph{arXiv}, 2024.

\bibitem{kumra2020antipodal}
S.~Kumra, S.~Joshi, and F.~Sahin, ``Antipodal robotic grasping using generative residual convolutional neural network,'' in \emph{IROS}, 2020.

\bibitem{zhou2022conditional}
K.~Zhou, J.~Yang, C.~C. Loy, and Z.~Liu, ``Conditional prompt learning for vision-language models,'' in \emph{CVPR}, 2022.

\bibitem{ainetter2021end}
S.~Ainetter \emph{et~al.}, ``End-to-end trainable deep neural network for robotic grasp detection and semantic segmentation from rgb,'' in \emph{ICRA}, 2021.

\bibitem{morrison2018closing}
D.~Morrison \emph{et~al.}, ``Closing the loop for robotic grasping: A real-time, generative grasp synthesis approach,'' \emph{arXiv}, 2018.

\bibitem{fang2020graspnet}
H.-S. Fang \emph{et~al.}, ``Graspnet-1billion: A large-scale benchmark for general object grasping,'' in \emph{CVPR}, 2020.

\bibitem{beck2023singularity}
F.~Beck \emph{et~al.}, ``Singularity avoidance with application to online trajectory optimization for serial manipulators,'' \emph{IFAC-PapersOnLine}, 2023.

\bibitem{vu2023machine}
M.~N. Vu, F.~Beck, M.~Schwegel, C.~Hartl-Nesic, A.~Nguyen, and A.~Kugi, ``Machine learning-based framework for optimally solving the analytical inverse kinematics for redundant manipulators,'' \emph{Mechatronics}, 2023.

\end{thebibliography}
